\documentclass[a4paper]{article}

\usepackage{INTERSPEECH2021}
\usepackage{url}

\title{Shrinking Bigfoot: Reducing wav2vec 2.0 footprint*}
\name{Zilun Peng$^1$ \thanks{*Submitted to INTERSPEECH 2021.}, Akshay Budhkar$^1$, Ilana Tuil$^2$, Jason Levy$^2$, Parinaz Sobhani$^1$, Raphael Cohen$^2$, Jumana Nassour$^2$}
\address{
  $^1$Georgian.io\\
  $^2$Chorus.ai}
\email{zilun.peng@georgian.io, akshay@georgian.io, ilana.tuil@chorus.ai, jason@chorus.ai, parinaz@georgian.io, raphael@chorus.ai, jumana.nassour@chorus.ai}

\begin{document}

\maketitle
\begin{abstract}
Wav2vec 2.0 is a state-of-the-art speech recognition model which maps speech audio waveforms into latent representations. The largest version of wav2vec 2.0 contains 317 million parameters. Hence, the inference latency of wav2vec 2.0 will be a bottleneck in production, leading to high costs and a significant environmental footprint. To improve wav2vec's applicability to a production setting, we explore multiple model compression methods borrowed from the domain of large language models. 
Using a teacher-student approach, we distilled the knowledge from the original wav2vec 2.0 model into a student model, which is 2 times faster and 4.8 times smaller than the original model. This increase in performance is accomplished with only a 7\% degradation in word error rate (WER). Our quantized model is 3.6 times smaller than the original model, with only a 0.1\% degradation in WER. To the best of our knowledge, this is the first work that compresses wav2vec 2.0.

\end{abstract}

\section{Introduction}

Speech recognition technologies enhance people's lives through a wide range of applications such as virtual assistants, home automation, and real-time transcription and captioning.  Neural network-based speech recognition models achieve superior performances, but successfully training them requires a lot of labeled data \cite{amodei2015deep}. 

Wav2vec 2.0 \cite{baevski2020wav2vec} is a framework that combines self-supervised masking pre-training with a tuning step that achieves impressive results with little labeled data. With only 10 minutes of labeled data, wav2vec 2.0 achieves word error rates (WER) of 4.8\% and 8.2 \% on LibriSpeech \cite{librispeech} dev-clean and dev-other datasets, respectively. Moreover, Wav2vec 2.0 achieves state-of-the-art performances on the LibriSpeech benchmark while using a minimal amount of labeled data. Self-supervised learning enables wav2vec 2.0 to learn speech representations efficiently without much labeled data. This allows easily training an end-to-end automatic speech recognition (ASR) system for a low resource language\footnote{For example, an Arabic ASR model https://huggingface.co/elgeish/wav2vec2-large-xlsr-53-arabic} or for a specific domain. The largest version of wav2vec 2.0 has 317 million parameters, which makes real-time inference inefficient in production. 

Language modeling is a central research question in natural language processing (NLP). In recent years BigLM architectures, based on novel pre-training approaches, revolutionized NLP \cite{devlin2019bert,howard2018universal}. The BERT \cite{devlin2019bert} architecture of masked pre-training and transformers is similar to wav2vec2.0. To improve wav2vec's applicability to a production setting, we explore multiple model compression methods borrowed from the domain of BigLMs. 

A significant drawback of BigLM models is high resource consumption. The negative impact of such models, beyond the costs for organizations using this technology in production, is cause for concern \cite{strubell2019energy, bender2021dangers}.  Various compression methods have been developed to overcome the problem, such as \cite{sanh2020distilbert}. To compress wav2vec 2.0, we explore quantization and knowledge distillation methods. Our quantized model is 3.6 times smaller than the original model, with a 0.1\% loss in WER, making it easy to deploy in a production environment. Our student model is at least 2 times faster on both CPU and GPU and 4.8 times smaller than the original model, with a 7\% loss in WER through knowledge distillation. 

Our work makes two key contributions: 1. This research is the first work that compresses the state-of-the-art wav2vec 2.0 speech recognition model to the best of our knowledge. 2. Our distilled student model has a faster inference speed and makes wav2vec 2.0 more cost-efficient and environmentally friendly. 

\section{Related work}
In this section, we relate research works done on model compression in ASR systems. Then we describe compression methods applied in language modeling, which we wish to explore for compressing wav2vec. 

\subsection{Model compression in ASR systems}
\label{sub_section:asr_compression}
Several works have been done on compressing end-to-end ASR systems to make them more applicable to production settings. The compression method varies depending on the ASR architecture used. Some of the methods result in approximately 2-3\% reduction in accuracy \cite{mehrotra2020iterative,li2019improving,mori2018compressing}, while others result in an improved accuracy \cite{winata2020lightweight,kurata2018improved}. 
Mehrotra \textit{et al.} \cite{mehrotra2020iterative} use reinforcement learning and low rank factorization to compress a system composed of an encoder-attention-decoder with unidirectional LSTMs, while Kurata \textit{et al.} \cite{kurata2018improved} and Takashima \textit{et al.} \cite{Takashima2018distilling} employ knowledge distillation to train a UniLSTM model from a BiLSTM one. Mori \textit{et al.} \cite{mori2018compressing} start with ``Deep Speech 2'' architecture, which consists of both convolution layers and bidirectional gated recurrent unit layers (GRU), reduce the number of parameters using Tensor-Train decomposition on the weight matrix of the GRUs (TT-GRU).

Some attempts have been made to compress transformer-based ASR systems, like the work done to share parameters across different layers by incorporating additional features related to the topic and the speaker \cite{li2019improving}, leading to less than a two-point decrease in accuracy. Another work proposed using a low-rank transformer (LRT) to reduce the number of parameters and boost the speed of training and inference \cite{winata2020lightweight}. The proposed LRT model improved accuracy without using a language model or acoustic data in addition to the improvement in size and speed.

\subsection{Classic compression methods in language modeling}

 Knowledge distillation, quantization, and pruning are few techniques employed in research for compressing language models.

\textbf{Knowledge distillation} involves the use of a teacher model (or models) to improve the performance of another model (student model). When used for compression, the student model's size is significantly smaller than that of the teacher model. 

Knowledge distillation in ASR is used both to improve accuracy \cite{futami2020distilling,gao2020distilling,mori2018compressing}, and to make the model more applicable to production by reducing its size \cite{kurata2018improved,Takashima2018distilling}, see subsection \ref{sub_section:asr_compression}.  Futami \textit{et al.}  \cite{futami2020distilling} use knowledge distillation to generate more syntactically or semantically likely results by providing a left context seq2seq ASR system (student model) with soft labels of context to the right. The authors produce soft labels by using a pre-trained BERT model (teacher model). Gao \textit{et al.} \cite{gao2020distilling} use multiple teacher models to train a student ASR model to improve ASR accuracy jointly. Moriya \textit{et al.} \cite{moriya2020distilling} improve the accuracy by adding another term to the loss function called self-distillation (SD), which comes from incorporating the teacher model. 

BERT \cite{devlin2019bert} is one of the most prominent language models in NLP. Like wav2vec, it is based on the transformer model, but it uses only encoders without decoders. DistilBERT \cite{sanh2020distilbert} is a distilled version of BERT, with a 40\% decrease in model size, a 60\% increase in speed, and a 97\% preservation of language understanding capabilities. Hence, it would be interesting to apply a similar distillation technique to wav2vec. 

\textbf{Quantization} involves reducing the number of bits used to represent weights. Normally, networks use 32-bit floating-point weights. Because neural networks have millions of such weights, quantizing these weights can make neural networks more energy efficient \cite{dally2015high}. Several works have been published on quantization, reducing weights to 8-bit \cite{8bitQuantize}, 4-bit \cite{choi2018pact}, 2-bit \cite{hwang2014fixed}, and 1-bit \cite{courbariaux2016binaryconnect} integers. Weights are static and can be quantized through scaling and shifting, but this is not true for activations. Activations vary depending on input samples and can be quantized through static or dynamic quantization. Static quantization learns scaling and shifting of activations through a batch of samples, while dynamic quantization scales and shifts activations in each forward pass. There are two popular software packages for quantization: Distiller \cite{nzmora2019distiller}, and PyTorch \cite{NEURIPS2019_9015}. Distiller simulates quantization; hence, weights are still represented using 32-bit floating-point numbers but are restricted to integer values. Since we are interested in studying the efficiency of quantizing wav2vec 2.0, we decided not to explore Distiller for quantization. In section \ref{sec:quantization}, we discuss the pros and cons of using PyTorch for quantizing wav2vec 2.0.

\textbf{Pruning} is executed by removing small weights that are likely redundant and can be removed without much loss in accuracy \cite{han2015learning}. Han \textit{et al.} \cite{han2016deep} proposed deep compression, an efficient compression pipeline, which first prunes the neural network model and then quantizes the remaining effective weights. We conducted a preliminary study on pruning using the Distiller \cite{nzmora2019distiller} sensitivity pruner. We set the sensitivity for each convolution layer to 0.1. We set the sensitivity in layers 3 to 16 of the transformer to 0.3 and layers 17 to 23 to 0.4. We pruned weights smaller than the product of the sensitivity and standard deviation of weights on that layer. The pruned model has a 23\% sparsity while maintaining the same WER as the original model. We, however, didn't observe any significant improvement in the inference speed.

\section{Wav2vec}
Wav2vec 2.0 \cite{baevski2020wav2vec} achieved a breakthrough in ASR by adopting the masked pre-training method employed in the massive language model BERT. BERT masks a few words in each training sentence, and the model learns by attempting to fill the gaps. Instead of masking words, wav2vec 2.0 masks parts of the audio representation and lets the transformer network fill in the gaps. This self-supervised setup makes it possible to learn mainly from unlabeled data while only fine-tuning on a small labeled dataset. The wav2vec 2.0 model comprises two significant components that we consider for compression: the CNN layers and the transformer layers.

In the following sections, we explore how compressing a fine-tuned version of wav2vec 2.0\footnote{We used wav2vec\_big\_960h as our base/original model. (https://dl.fbaipublicfiles.com/fairseq/wav2vec/wav2vec\_big\_960h.pt)} using knowledge distillation and quantization affects the size of the model and its inference speed.

\section{Knowledge distillation for wav2vec 2.0}
We based our distillation on the knowledge distillation used by Sanh \textit{et al.} to compress BERT \cite{devlin2019bert} to DistilBERT \cite{sanh2020distilbert}. They initialized their student models by taking alternating layers from a well-performing larger model and scheduled the learning rate by first increasing, then gradually decreasing it. 

Since it is possible to reduce either the CNN layer or the transformer layers of wav2vec 2.0, we decided to reduce the transformer ones since wav2vec 2.0 has only 7 CNN layers, which does not constitute a substantial computational bottleneck compared to transformer layers, see Table \ref{tab:fewer_conv_layers}. 




\subsection{Knowledge distillation loss}
The classic knowledge distillation approach \cite{hinton2015distilling} focuses on transferring knowledge between classification neural networks, where both the teacher and the student models output probability distributions over $M$ labels. In our case, the output is latent representations of the input speech audio waveform. We project these representations into probability distributions over $M$ tokens, where a token could map to a character or word boundary. Let $T$ and $S$ be the output probability distributions of the teacher and the student model, respectively. Inspired by \cite{hinton2015distilling}, we define the knowledge distillation loss to be the Kullback–Leibler (KL) divergence between $T$ and $S$:

\begin{equation}
\label{eq:kd_loss}
    L_{distill} (T, S) = \frac{1}{N}\sum_{0 \leq i < N}\sum_{0 \leq j < M} T_{ij} \log \left( \frac{S_{ij}}{T_{ij}} \right)
\end{equation}

The dimension of $T$ and $S$ is $N \times M$, where $N$ is the length of the representation and $M$ is the number of tokens. The KL divergence signals how far off is the student model's prediction is from its teacher, and the student model learns from the teacher by optimizing this loss.

\subsection{The architecture of the student model}
\label{sec:student_model_architecture}
The only difference between our student model and the original wav2vec 2.0 model is the number of transformer layers. The largest version of wav2vec 2.0 has seven convolution layers and 24 transformer layers, while our student models have a reduced number of transformer layers. Our smallest student model has four transformer layers. We have also trained student models with 6, 8, 10, 12 transformer layers. See section \ref{sec:exp} for a discussion on the trade-off between the number of transformer layers and the performance of the different student models.

\subsection{Objective function}
Our objective function for knowledge distillation training consists of the knowledge distillation loss and feature penalty \cite{baevski2020wav2vec}, $L_{distill} + L_{feature}$. $L_{distill}$, the knowledge distillation loss, is defined in equation \ref{eq:kd_loss}, and $L_{feature}$ is the L2 norm of the final output of convolution layers and serves as a regularization term for convolution layers. The original wav2vec 2.0 uses $L_{feature}$ during training. 

We have tried to use the cosine embedding loss \cite{sanh2020distilbert} in our objective function, but it didn't help with the student model's performances.

\subsection{Training techniques}
After trying different optimizers, step size scheduling, objective functions, and other techniques to help the student model learn faster, the two approaches that significantly improved the student model's performance were using DistilBERT's initialization strategy and training on more data.

\section{Quantization  for wav2vec 2.0}
\label{sec:quantization}
In this section, we discuss how the two quantization techniques available in Pytorch apply to our task of quantizing wav2vec 2.0.

\subsection{Static quantization}
Static quantization quantizes weights and requires calibration before inference to determine scaling and shifting for quantizing activations. PyTorch does not support quantization for gelu activation \cite{hendrycks2020gaussian} (as of March 2021), one of the activations used in the wav2vec 2.0 model. A closed pull request \footnote{https://github.com/pytorch/pytorch/pull/34396} shows they were trying to add the support but suspended it temporarily due to some test case failures. One possible workaround is to dequantize before gelu and re-quantize afterward, but this causes some performance degradation. 

\subsection{Dynamic quantization}
\label{sec:dynamic_quantization}
 Dynamic quantization only quantizes weights and dynamically quantizes activations in each forward pass. As of March 2021, there are no supports for dynamically quantizing the multi-head attention (MHA) mechanism, which is an essential part of a transformer layer. Related issues remain open in PyTorch \footnote{https://github.com/pytorch/pytorch/issues/32764} and fairseq \footnote{https://github.com/pytorch/fairseq/issues/1901}. As a result, tensors entering MHA must not be quantized. We observed that dynamic quantization wraps weights and biases of MHA with some methods, requiring a costly  \textit{linear\_unpack} operation to access them during inference. We save weights and biases before inference so that \textit{linear\_unpack} is not performed for every sample during inference, hence increasing inference efficiency.



\section{Experiments}
The wav2vec 2.0 model we are trying to compress was trained with 960 hours of unlabeled data from the LibriSpeech dataset \cite{librispeech}, and then fine-tuned with the labeled version of the same 960 hours. We did not attempt to compress the model before it was fine-tuned. We used a Viterbi decoder to convert the output of the student wav2vec 2.0 model into text. We conducted training on 4 Nvidia V100 GPUs and set the batch size to 1 per GPU. We trained with 16-bit floating-point arithmetic and fixed the seed at 42 in all our runs to make the results reproducible.

For knowledge distillation, we used the Adam optimizer with weight decay. We set $\text{eps}=10^{-6}$, $\text{betas}=(0.9, 0.98)$, and the weight decay coefficient to 0.01. We scheduled the learning rate by the number of epochs. The learning rate starts from $2.5\times 10^{-5}$ and linearly increases for the first two epochs. After two epochs, we linearly decrease the learning rate in every epoch. Also, we assessed the effect of using a different number of transformer layers on the target metrics, see Table \ref{tab:trade_off}.

For quantization, we used PyTorch's dynamic quantization to quantize linear layers of wav2vec 2.0. 
Table \ref{tab:compression_results} contains the results of evaluating the different compression methods on LibriSpeech train-clean-100. 

Our implementation is available at \url{https://git.io/JmpJM}.





\section{Results and discussion}
\label{sec:exp}

\begin{table*}
\centering
\begin{tabular}{|l|l|l|l|l|l|}
\hline
\textbf{Model} & \textbf{\#Parameters} &  \textbf{Model Size} & \textbf{CPU Inference Time} &   \textbf{GPU Inference Time}  & \textbf{WER} \\
\hline
Original & 317 M & 1262 mb &  4433 s & 123 s  & \textbf{2.63\%} \\
\hline
Distilled &  \textbf{65 M} &\textbf{ 262 mb} & \textbf{1560 s} & \textbf{51 s}  & 9.51\%\\
\hline
Quantized & 317 M & 354 mb & 4079 s& N/A  & 2.75\%\\
\hline 
\end{tabular}
\caption{Performances of the original wav2vec 2.0 model, a student model trained using knowledge distillation, and a quantized wav2vec 2.0 model.}
\label{tab:compression_results}
\end{table*}

\begin{table}
\centering
\begin{tabular}{|l|l|l|l|l|}
\hline
 \textbf{Layers}& \textbf{Parameters} & \textbf{Model} & \textbf{Inference} & \textbf{WER} \\
 &  & \textbf{Size} & \textbf{Time} &  \\
\hline
24 & 317 M & 1262 mb & 123 s & 2.63\% \\
\hline
12 & 166 M & 665 mb & 81 s & 6.6\% \\
\hline
10 & 141 M & 564 mb & 75 s & 7.59\% \\ 
\hline
8 & 115 M & 464 mb & 69 s & 10.42\% \\ 
\hline
6 & 91 M & 363 mb & 60 s & 16.54\% \\ 
\hline
\end{tabular}
\caption{Performance of a student wav2vec 2.0 model with different number of transformer layers on GPU.}
\label{tab:trade_off}
\end{table}

\begin{table}[!htbp]
\centering
\begin{tabular}{|l|l|l|}
\hline
\textbf{Layers} & \textbf{GPU Inference Time}  &  \textbf{CPU Inference Time} \\
\hline
7  & 2.97 s  & \textbf{161 s} \\
\hline
4  & \textbf{ 2.26 s } & 387 s \\
\hline
\end{tabular}
\caption{Inference time of wav2vec 2.0 model with different number of convolution layers. Original wav2vec 2.0 model has 7 convolution layers.}
\label{tab:fewer_conv_layers}
\end{table}

Table \ref{tab:compression_results} shows the performance of the largest version of wav2vec 2.0 compared to the compressed models. The distilled model has a 7\% loss in terms of word error rate (WER) compared with the original model. However, the distilled model is at least two times faster than the original model on both CPU and GPU, in addition to being  4.8 times smaller than the original wav2vec 2.0 model. We trained the distilled model for 32 epochs.

Considering the quantized model, we notice that even though the model is 3.6 times smaller than the original model, it is not significantly faster. This discrepancy can be attributed to the need for dequantize tensors entering multi-head attention; see section \ref{sec:dynamic_quantization} for more details. PyTorch does not support GPU quantization; therefore, we report only CPU inference time for the quantized model. Its significant reduction in size, together with its ability to maintain the original WER, albeit with a slight loss, make the quantized model easily deployable in a production environment.


\subsection{Knowledge distillation using different settings}
There is a wide range of settings that can be tweaked when distilling a model. We explore some of them to understand their effect on the resulting model and the target metrics.

We conducted some experiments to evaluate the trade-off between the size of a model and its performance when performing knowledge distillation by varying the number of kept transformer layers in the student model. LibriSpeech train-clean-100 was used to train the student models. We trained each student model for ten epochs. We performed inferences on a single GPU. Table \ref{tab:trade_off} reports the results of these experiments. The performance of the original wav2vec 2.0 is shown in the first row. We observe that as the number of transformer layers decreases, the inference time decreases, but WER increases. Thus, as expected, as the model becomes smaller, its inference time improves, but its accuracy degrades.

In section \ref{sec:student_model_architecture}, we introduced the student wav2vec 2.0 model's architecture, where we mentioned that reducing the number of convolution layers will probably not have much effect on the inference time of the distilled model. We tested this claim by reducing the convolution layers to 4. Table \ref{tab:fewer_conv_layers} shows the result of this experiment compared to the original setting. GPU inference time is only 0.7 seconds faster in the four convolution layer model, whereas CPU inference time is slower. This behavior is because we replaced convolution layers with max-pooling, and PyTorch has a slower max-pooling operation on the CPU. 

\begin{figure}
\vspace{-1.78em}
    \includegraphics[width=\linewidth]{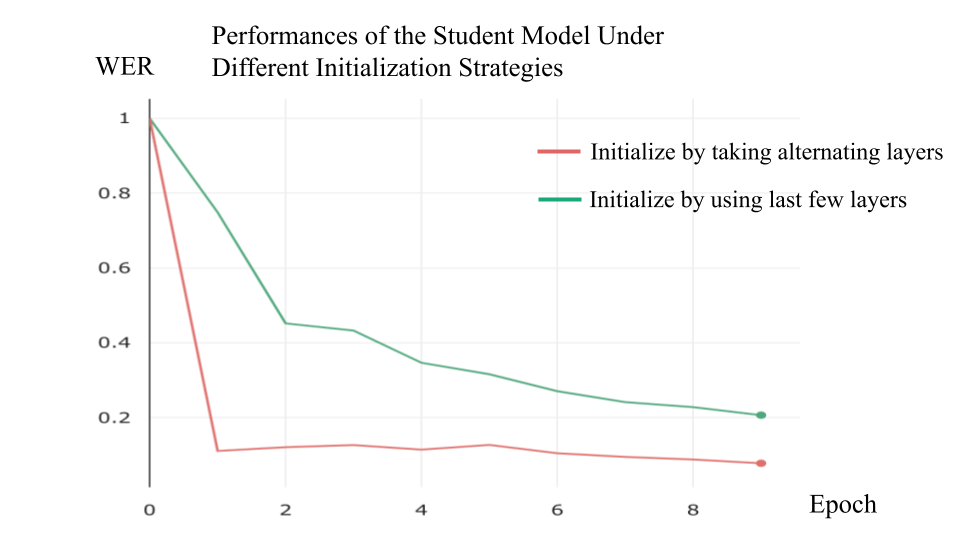}
    \caption{A student wav2vec 2.0 model initialized using different strategies, then trained with knowledge distillation. }
    \label{fig:init_strategy}

\end{figure}

DistilBERT initializes layers in the student model by taking alternating layers from a larger model. It's possible to initialize layers in the student model by taking the last few layers from a larger model. Figure \ref{fig:init_strategy} shows the effectiveness of DistilBERT's initialization strategy, as the student wav2vec 2.0 model initialized using DistilBERT's approach significantly outperforms the other method from the beginning of training. We split the dev-clean dataset into two parts for this experiment, one for training and the other for validation, which we used for reporting WER. 

\begin{figure}
\vspace{-0.78em}
    \includegraphics[width=\linewidth]{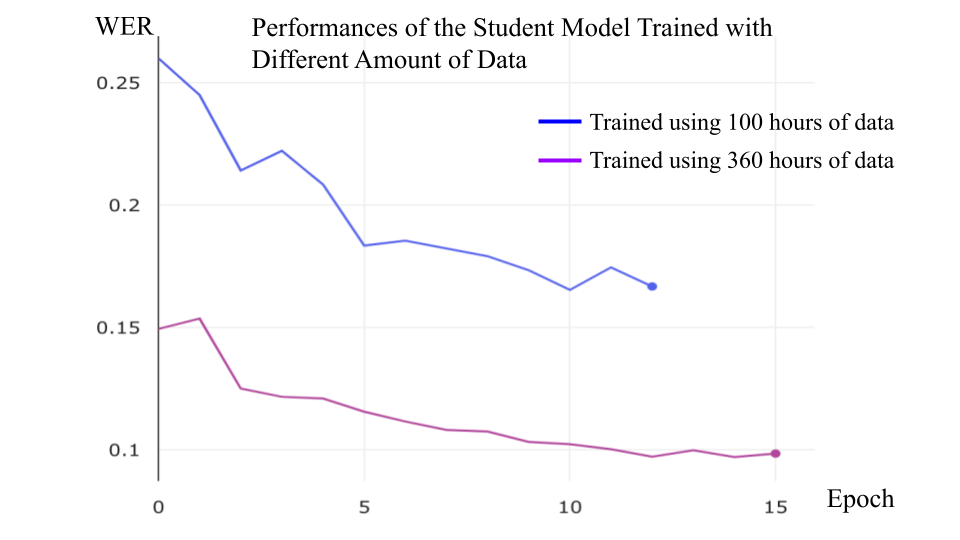}
    \caption{A student wav2vec 2.0 model trained with knowledge distillation using different amount of data.}
    \label{fig:train_data_effect}
\end{figure}

We observed that a student model trained with more data tends to obtain a better WER. As figure \ref{fig:train_data_effect} shows, the student model trained using 360 hours of data (train-clean-360) achieves better WER than the student model trained with less data (train-clean-100). 

\section{Conclusion}
Wav2vec models present a valuable technology for democratizing ASR. We explored multiple methods for reducing memory usage and speeding up such models for deployment. 

Quantizing reduced the memory footprint by 72\%; this is significant for reducing the costs of the machines running such models. We found that knowledge distillation is the most effective compression method for wav2vec 2.0 for saving energy by lowering CPU or GPU inference time. Using knowledge distillation, we trained a student wav2vec 2.0 model, which is at least two times faster than the original model on both CPU and GPU. As demonstrated by our experiments, we think that there is a lot of room to further improve the performance of a compressed model by using more data in knowledge distillation training.

\bibliographystyle{IEEEtran}
\bibliography{template}

\end{document}